\documentclass{article}
\usepackage{spconf,epsfig}
\usepackage{amsmath,amsfonts}
\usepackage{algorithmic}
\usepackage{caption}
\usepackage{algorithm}
\usepackage{array}
\usepackage{fancyhdr}
\usepackage{enumitem}
\usepackage[caption=false,font=normalsize,labelfont=sf,textfont=sf]{subfig}
\usepackage{textcomp}
\usepackage{stfloats}
\usepackage{url}
\usepackage{verbatim}
\usepackage{graphicx}
\usepackage{cite}
\usepackage{multirow, booktabs}
\usepackage{amsthm}

\let\OLDthebibliography\thebibliography
\renewcommand\thebibliography[1]{
  \OLDthebibliography{#1}
  \setlength{\parskip}{0pt}
  \setlength{\itemsep}{0pt plus 0.3ex}
}

\begin{document}\sloppy

% Example definitions.
% --------------------
\def\x{{\mathbf x}}
\def\L{{\cal L}}

% Title.
% ------
\title{Open Set Domain Adaptation By Novel Class Discovery}
%
% Single address.
% ---------------
\name{Jingyu Zhuang, Ziliang Chen, Pengxu Wei, Guanbin Li, Liang Lin}
%Address and e-mail should NOT be added in the submission paper. They should be present only in the camera ready paper. 
\address{Sun Yat-sen University \\
zhuangjy6@mail2.sysu.edu.cn, c.ziliang@yahoo.com,\\ \{weipx3,liguanbin\}@mail.sysu.edu.cn, linliang@ieee.org}

\maketitle
\thispagestyle{fancy}         %更改plain状态
\fancyhead{}                     %清除以前的命令
\lhead{Published as a conference paper at ICME 2022} 
\lfoot{}
\cfoot{\thepage}  %current page number
\rfoot{} 

\pagestyle{plain}
\cfoot{\thepage}

\renewcommand{\headrulewidth}{0pt} 

\begin{abstract}
In \textit{Open Set Domain Adaptation} (\textbf{OSDA}), large amounts of target samples are drawn from the implicit categories that never appear in the source domain. Due to the lack of their specific belonging, existing methods indiscriminately regard them as a single class ``\emph{unknown}''. We challenge this broadly-adopted practice that may arouse unexpected detrimental effects because the decision boundaries between the implicit categories have been fully ignored. Instead, we propose Self-supervised Class-Discovering Adapter (\textbf{SCDA}) that attempts to achieve OSDA by gradually discovering those implicit classes, then incorporating them to restructure the classifier and update the domain-adaptive features iteratively. SCDA performs two alternate steps to achieve implicit class discovery and self-supervised OSDA, respectively. By jointly optimizing for two tasks, SCDA achieves the state-of-the-art in OSDA and shows a competitive performance to unearth the implicit target classes. 

\end{abstract}


\section{Introduction}
\label{sec:intro}

Unsupervised Domain Adaptation (\textbf{UDA}) methods aim to transfer the knowledge from a labeled source domain to classify unlabeled samples in a target domain by minimizing the cross-domain distribution discrepancy.
Despite their impressive progress, UDA methods usually operate under the \emph{close-set} assumption, \emph{i.e.}, the source categories and the target categories should be exactly identical. 
However, this assumption is probably violated in the real world since target samples are collected from diverse classes even beyond source categories.
Therefore, \emph{Open Set Domain Adaptation} (\textbf{OSDA}) \cite{Busto2017Open,Saito2018Open} has attracted increasing attention, where the target domain may contain \textbf{\emph{implicit classes}} that never appear in the source domain.

\begin{figure}[t]
  \centering
  \includegraphics[width=0.9 \columnwidth]{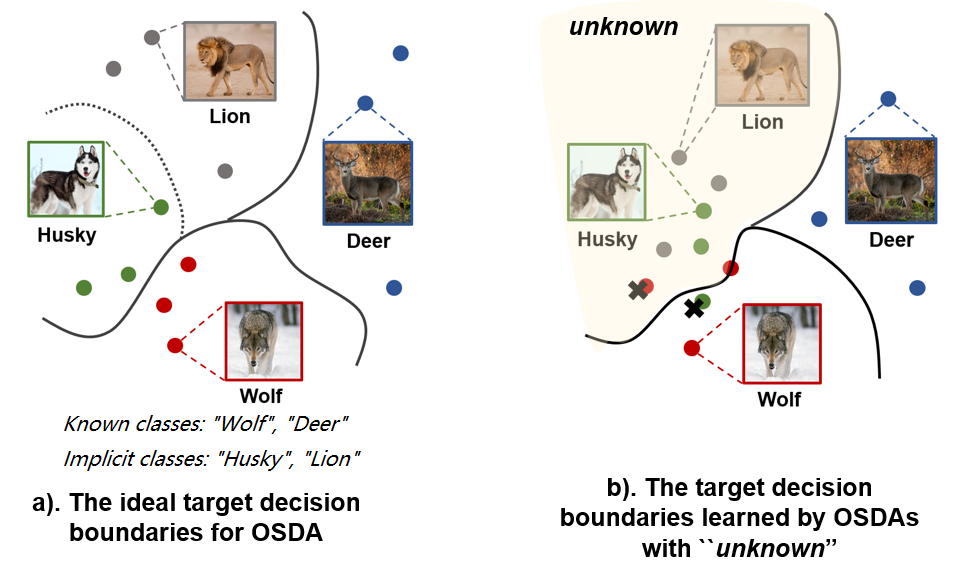}
   \caption{Existing OSDA methods unitedly treat implicit classes as a single negative class ``\emph{unknown}''. Our work challenges this broadly-accepted practice, since ignoring decision boundaries between implicit classes results in all implicit-class features converging together due to the cluster assumption, which further causes a detrimental effect.  For instance, suppose \emph{husky}, \emph{lion} denote the implicit classes and \emph{wolf}, \emph{deer} denote the known classes. Given a known class similar with an implicit class, \emph{e.g.}, \emph{wolf} v.s. \emph{husky}, the \emph{wolf} target features could be attracted to the feature field of \emph{lion} since \emph{husky} and \emph{wolf} features are hard to distinguish, yet \emph{husky} and \emph{lion} tend to converge to the identical center in the ``\emph{unknown}''. } 
\label{pic_01}
\end{figure}

Due to the absence of both the corresponding categories for target samples and the number for implicit classes, existing OSDA methods regard all target samples of implicit classes as a single class ``\emph{unknown}''.
This practice is straightforward but probably problematic. 
Specifically, existing OSDA methods \cite{liang2020we, kundu2020towards, xu2020joint} aim to minimize the cross-entropy losses of known classes and the ``\emph{unknown}'' class. 
Under the cluster assumption \cite{shu2018dirt}, the features of unknown target samples are optimized to converge to an identical center due to the shared labels.
Whereas, since their intrinsic structure and diversity have been ignored, the convergence is hard to achieve in practice. Especially when unknown target samples contain more categories, the features of unknown samples will be more probably mixed up with the known-class features around the decision boundaries between known classes and the ``\emph{unknown}'' class (Figure.\ref{pic_01}.b), hence, breeding the potential performance drop.

In this paper, we focus on a new methodology to achieve OSDA from another point of view. Instead of fabricating the ``\emph{unknown}'', we aim to transfer the source knowledge along with discovering implicit classes \cite{han2020automatically} in unknown target samples. The process automatically estimates the number of implicit classes and how they are distributed, then leverages this self-supervised information to progressively update the OSDA model to improve the performance of OSDA.

%, which simultaneously achieves adaptation and implicit class discovery.

\begin{figure*}[t]
  \centering
  \includegraphics[width=0.8 \textwidth]{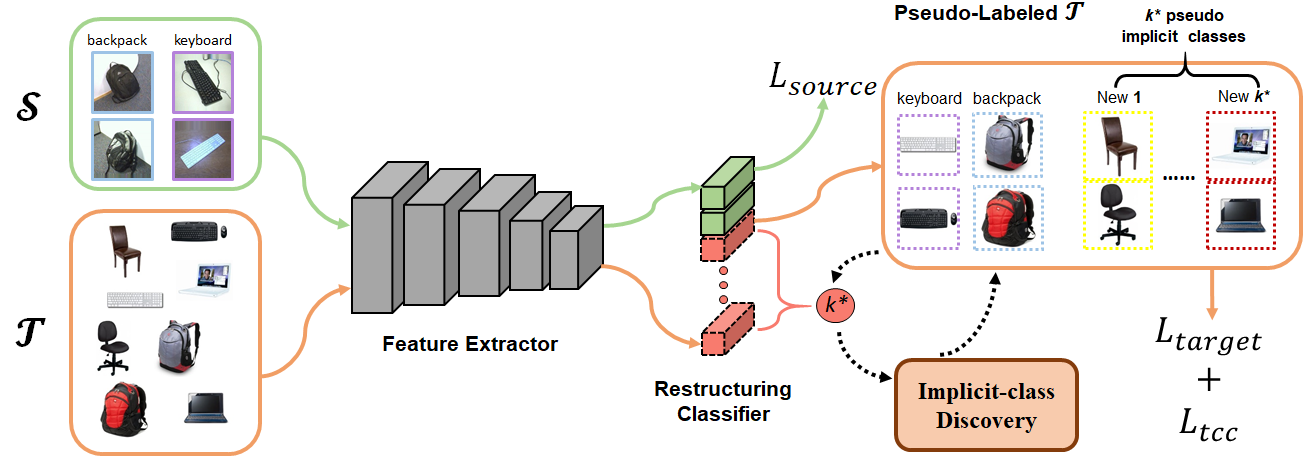}
  \caption{An overview of SCDA. SCDA starts from a pre-training to obtain unknown target samples. Then SCDA designs an algorithm to discover the implicit classes in the unknown target samples; Afterwards, SCDA restructures the classifier and updates it to recognize them.}
  
  \label{pic_02}     % Give a unique label

\end{figure*}

To \hspace{-0.2em}this \hspace{-0.2em}end, \hspace{-0.2em}we \hspace{-0.2em}propose \hspace{-0.2em}Self-supervised \hspace{-0.2em}Class-Discovering Adapter (SCDA). After an adversarial pre-training, SCDA separates unknown target samples roughly. 
Unlike previous OSDA methods that regard all unknown target samples as the ``\emph{unknown}'' and neglect their inter-class structure, SCDA designs an unsupervised algorithm, \emph{i.e.}, implicit class discovery, to estimate the number of the implicit classes and group the unknown samples belonging to each implicit class.
Based on the results of implicit class discovery, SCDA restructures and updates the model to push features of different classes away from each other to reduce the risk of confusing them.
Our contributions are summarized in two aspects:
\begin{itemize}[leftmargin=*]
	\item We consider OSDA from a new point of view: Instead of regarding target samples beyond categories in the source domains as the ``\emph{unknown}'' class, we attempt to discover their structure along with the domain adaption.
  \item We propose Self-supervised Class-Discovering Adapter (SCDA) which combines discovering the implicit classes and learning domain-invariant features in a framework. 
	\item Extensive experiments on three OSDA benchmarks are conducted to evaluate SCDA. The results evidence the superiority of SCDA on both OSDA and implicit class discovery.
\end{itemize}

\section{Related Work}

\textbf{Close-set domain adaptation (UDA).} 
Assuming labeled source and the unlabeled target domains share their label sets, closed-set UDA methods aim at reducing the domain shift across the source and target domains. Most existing algorithms are based on either domain discrepancy matching \cite{long2015learning, long2017deep} or adversarial learning \cite{Yaroslav2017Domain}. The former employs various measurements to minimize the distribution discrepancy; The latter adversarially learns domain invariant features.

\noindent \textbf{Open-set domain adaptation (OSDA).} 
OSDA learners categorize target samples into the known classes or \emph{unknown}. OSBP \cite{Saito2018Open} learns representations to separate unknown target samples through an adversarial method. STA \cite{liu2019separate} trains a binary classifier to perform fine separation on all target samples and weight each target sample to alleviate negative-transfer caused by unknown target samples. Besides, there are a few works, \emph{e.g.}, TIM \cite{kundu2020towards}, SHOT \cite{liang2020we}, JPOT \cite{xu2020joint} and PGL \cite{luo2020progressive}. However, all these approaches simply assign target samples beyond the label-set of source classes to the \textit{unknown}, while our method further classifies each novel class inside them.

\noindent \textbf{Novel-class discovery.} 
Provided the labeled data from related but different classes, novel-class discovery aims to find novel classes in unlabeled data. However, existing methods, \emph{e.g.}, DTC \cite{han2019learning} and \cite{han2020automatically}, assume that the labeled and unlabeled data are drawn from identical distribution and non-overlapping classes. They can not be directly adopted for OSDA, since the domain shift between the labeled and unobserved data would incur a severe performance degeneration.

\section{Methodology}

In this section, we first define the problem setting of OSDA; then we introduce the architecture of our SCDA; finally, we describe the training process in detail.

\subsection{Problem Setting} 

% \noindent \textbf{Problem Setting.} 

Suppose $n_{s}$ labeled images $\mathcal{S}={(\boldsymbol{x}_{i}^{s},\boldsymbol{y}_{i}^{s})}_{i=1}^{n_{s}}$ are drawn from a source density $P_{\mathcal{S},\mathcal{C}_{\mathcal{S}}}(\boldsymbol{x},\boldsymbol{y})$ and $n_{t}$ unlabeled images $\mathcal{T}={(\boldsymbol{x}_{i}^{t})}_{i=1}^{n_{t}}$ are drawn from a target density $P_{\mathcal{T},\mathcal{C}_{\mathcal{S}}}(\boldsymbol{x})=\int P_{\mathcal{T},\mathcal{C}_{\mathcal{S}}}(\boldsymbol{x}, \boldsymbol{y})d\boldsymbol{y}$. $\mathcal{C}_{\mathcal{S}}$ denotes the set of the source classes, $\mathcal{C}_{\mathcal{T}}$ denotes that of the target, and $\mathcal{C}_{\mathcal{T}}/\mathcal{C}_{\mathcal{S}}$ denotes the implicit classes in $\mathcal{T}$. In OSDA, due to $\mathcal{C}_{\mathcal{T}}/\mathcal{C}_{\mathcal{S}}\neq \emptyset$, we are required to classify target samples of $|\mathcal{C}_{\mathcal{S}}|$ known classes correctly ($|A|$ indicates the number of members in $A$) and simultaneously reject the unknown target samples belonging to $\mathcal{C}_{\mathcal{T}}/\mathcal{C}_{\mathcal{S}}$.

\subsection{Overall Architecture} 

SCDA can be flexibly deployed to existing neural network architectures. As shown in Figure.\ref{pic_02}, 
the architecture of SCDA consists of two modules: a \emph{feature extractor} $F$ proposed to learn class-aware domain-invariant features; a \emph{dynamically restructuring classifier} $C$ proposed to classify unlabeled target samples into the known classes $\mathcal{C}_{\mathcal{S}}$ and the implicit classes $\mathcal{C}_{\mathcal{T}}$.
The output dimension of $C$ is initialized as $|\mathcal{C}_{\boldsymbol{S}}|+1$, where the $(|\mathcal{C}_{\boldsymbol{S}}|+1)'$-th class indicates the unknown target samples. It is worth noting that, according to the results of implicit class discovery, the output dimension of $C$ alters to $|\mathcal{C}_{\mathcal{S}}|+k_t^\ast$, where $k*$ refers to the newly discovered classes.

\subsection{Self-supervised Class-Discovering Adapter}

The pipeline of SCDA mainly consists of two alternate steps, \emph{i.e.}, implicit class discovery (Sec.\ref{step_1_dis}) and self-supervised OSDA (Sec.\ref{step_2_dis}).
Briefly, SCDA employs a pre-training to roughly separate the unknown target samples. Then, it alternately performs two steps to discover the implicit classes in $\mathcal{C}_{\mathcal{T}}/\mathcal{C}_{\mathcal{S}}$ and further improve the performance of OSDA based on the results of discovery.
First, SCDA discovers the implicit classes in $\mathcal{C}_{\mathcal{T}}/\mathcal{C}_{\mathcal{S}}$ by estimating their number and constructing the pseudo implicit classes through a clustering assignment. 
Then based on the pseudo implicit classes, SCDA restructures the $C$ to recognize newly discovered classes; SCDA trains $C$ along with $F$ to diminish the domain gap so that the model can be generalized to classify the target samples in implicit classes.
SCDA repeatedly executes two alternate steps until the maximal epoch is reached.
We elaborate the pre-training and the two steps in the following three subsections. The pipeline of SCDA is found in Supplementary Algorithm.2.

\subsubsection{\textbf{Pre-training}}\label{step_0}

Due to the absence of target samples' labels, SCDA utilizes adversarial learning in this pre-training step to preliminarily separate the unknown target samples.
In brief, $C$ is trained to confuse the known and unknown target samples while $F$ is trained oppositely to distinguish them. We utilize the correlation confusion derived from \cite{jin2020minimum} to implement the adversarial training. Specifically, given a mini-batch of $m$ target samples, each element in a class correlation matrix $\mathbf{R}$ presents as:
\begin{equation}\label{R}
	\mathbf{R}_{i,j}\hspace{-0.3em}=
	\mathbf{\hat{y}}^\top_{i,\cdot}\frac{m\Big(1+\exp\big(-H(\boldsymbol{x}_{i}^{(t)};F,C)\big)\Big)}{\sum^{m}_{i'=1}\Big(1+\exp\big(-H(\boldsymbol{x}_{i'}^{(t)};F,C)\big)\Big)}\mathbf{\hat{y}}_{j,\cdot}
\end{equation}
where $\mathbf{\hat{y}}_{j,\cdot}$ represents the softmax output for the class-$j$ prediction \hspace{-0.1em}over $m$ target examples in the mini-batch and $H$ (Eq.\ref{entropy}) is a measure to increase the weight of the reliable examples.

The $j'$-th column in $\mathbf{R}$ measures the correlation between the $j'$-th class and other classes when $C$ classifies a mini-batch of samples. The higher $\mathbf{R}_{j,{j}'}$ implies that $C$ will more probably classify the samples drawn from the $j$-th class to the $j'$-th class. So we can adjust the value of $\mathbf{R}$ between known and unknown classes to cause confusion.
To be specific, after normalizing $\mathbf{R}$ by the sum of each row, we obtain $\mathbf{\hat{R}}$ where the summation over each row is 1.
Then we optimize the value of $\mathbf{\hat{R}}_{j,|\mathcal{C}_{\boldsymbol{S}}|+1}$ to 0.5 which means the probability that $C$ classifies samples into the unknown class or ${j}'$-th known class is equal, \emph{i.e.}, $C$ can not distinguish the known and unknown samples. While, we train $F$ in the opposite direction by inserting a reversed gradient layer \cite{Yaroslav2017Domain} between $C$ and $F$. The adversarial learning loss is defined as:
\begin{equation}
	L_{adv} = \mathbb{E}_{(\boldsymbol{x})\sim\mathcal{T}} L_{bce}( \frac{1}{|\mathcal{C}_{\boldsymbol{S}}|}\sum_{j=1}^{|\mathcal{C}_{\boldsymbol{S}}|} \mathbf{\hat{R}}_{j,|\mathcal{C}_{\boldsymbol{S}}|+1}, \frac{1}{2})
	\label{loss_adv}
\end{equation} 

Simultaneously, provided with source labeled data $\mathcal{S}$, we have a standard cross-entropy loss $L_{\rm s}$ to correctly categorize the known classes in $\mathcal{C}_{\mathcal{S}}$:
\begin{equation}
	L_{\rm s} = - \ \mathbb{E}_{(\boldsymbol{x},\boldsymbol{y})\sim\mathcal{S}} \ \boldsymbol{y}^{T}\log \ C\big(F(\boldsymbol{x})\big)
	\label{L_source}
\end{equation}

Besides, to alleviate the cross-known-class confusion caused by domain shift, we optimize $L_{\rm kcc}$ (Eq.\ref{L_kcc}). It is worth noting that, $L_{\rm kcc}$ does not punish the confusion to $\mathcal{C}_{\mathcal{T}}/\mathcal{C}_{\mathcal{S}}$, leading to the cross-domain features only aligned on $\mathcal{C}_{\mathcal{S}}$.

\begin{equation}
	L_{kcc} =  \mathbb{E}_{(\boldsymbol{x})\sim\mathcal{T}} \ \frac{1}{|\mathcal{C}_{\boldsymbol{S}}|}\sum_{j=1}^{|\mathcal{C}_{\boldsymbol{S}}|}\sum_{{j}'\neq j}^{|\mathcal{C}_{\boldsymbol{S}}|}\mathbf{\hat{R}}_{j,{j}'}
	\label{L_kcc}
\end{equation}	

Combing the above items, the overall pre-train objective is formulated as:
\begin{equation}
	\mathop{}_{F}^{\min}  L_{\rm s}-L_{\rm adv}+L_{\rm kcc}
	\label{step_1_F}
\end{equation}
\begin{equation}
	\mathop{}_{C}^{\min}  L_{\rm s}+L_{\rm adv}+L_{\rm kcc}
	\label{step_1_C}
\end{equation}

After preparation, SCDA alternately runs two steps to achieve implicit class discovery and self-supervised OSDA.

\begin{algorithm}[t]	
  \setlength{\itemsep}{-1.0cm}	
  \caption{Implicit Class Discovery in Sec.\ref{step_1_dis}} \label{Estimation_k}	
  {\bf Input:} Target dataset $\mathcal{T}$; pre-trained feature extractor $F$ and classifier ${C}$; max implicit classes number $k_{max}$.\\
  {\bf Output:} The estimation number $k^\ast$ of implicit classes; pseudo-labeled known target data $\hat{\mathcal{T}}_{\rm kn}$; pseudo-labeled newly discovered target data $\{\hat{\mathcal{T}}_{i}\}^{k^\ast}_{i=1}$.
  	
  \begin{algorithmic}[1]	
    \STATE Compute the entropy for $\boldsymbol{x}^{t}\sim \mathcal{T}$ by Eq. \ref{entropy}. Sort $\boldsymbol{x}^{t}$ based on their entropies. Select samples according to the entropy to build $\hat{\mathcal{T}}_{\rm kn}$ with pseudo labels $\hat{\boldsymbol{y}_{t}}$ and $\hat{\mathcal{T}}_{\rm im}$.
    \STATE Extract feature of $\hat{\mathcal{T}}_{\rm kn}$ and $\hat{\mathcal{T}}_{\rm im}$ using $F$.
    \STATE \textbf{For} $0 \leq k \leq k_{max}$ \textbf{do}
    \STATE \ Run k-means++ on the extracted feature with $k$ clusters.
    \STATE \ Compute CA for $\hat{\mathcal{T}}_{\rm kn}$ and SSE for $\hat{\mathcal{T}}_{\rm kn}\cup\hat{\mathcal{T}}_{\rm im}$.
    \STATE \textbf{End for}
    \STATE Let $\hat{k}$=$(k^\ast_{\rm CA}+k^\ast_{\rm elbow})/2$. $k^\ast_{\rm CA}$ is the value of $k$ maximizes CA. $k^\ast_{\rm elbow}$ is generated by the elbow method.
    \STATE Let $k^\ast$=$\hat{k}-|\mathcal{C}_{\mathcal{S}}|$. Run k-means++ on the features of $\hat{\mathcal{T}}_{\rm im}$ to obtain $k^\ast$ clusters $\{\hat{\mathcal{T}}_{i}\}^{k^\ast}_{i=1}$. Categorize $\hat{\mathcal{T}}_{\rm im}$ into $\{\hat{\mathcal{T}}_{i}\}^{k^\ast}_{i=1}$ with pseudo labels.
    \STATE \textbf{Return} $k^\ast$=$|\mathcal{C}_\mathcal{O}|$; $\hat{\mathcal{T}}_{\rm kn}$; $\{\hat{\mathcal{T}}_{i}\}^{k^\ast}_{i=1}$
  \end{algorithmic}
\end{algorithm}

\subsubsection{\textbf{Implicit class discovery}}\label{step_1_dis}

In this step, SCDA attempts to determine the number of implicit classes in $\mathcal{T}$ with the help of the labeled data. 
However, if we directly use the labeled source data, the domain shift between source and target domain would affect the accuracy of the estimation.
Hence, SCDA first constructs two \emph{\textbf{high-confident target candidates}} sets $\hat{\mathcal{T}}_{\rm kn}$ and $\hat{\mathcal{T}}_{\rm im}$ with pseudo labels, indicating target samples in known classes and implicit classes, respectively.
Then SCDA estimates $|\mathcal{C}_{\mathcal{T}}/\mathcal{C}_{\mathcal{S}}|$ by evaluating the \emph{\textbf{clustering consistency}} between $\hat{\mathcal{T}}_{\rm kn}$ and $\hat{\mathcal{T}}_{\rm im}$, and assigns pseudo labels to the newly discovered classes.
This step has been summarized in Algorithm.\ref{Estimation_k}.

\textbf{High-confident target candidates.} 
Instead of analyzing whole $\mathcal{T}$, we select target candidates with higher cross-domain classification consistency, because the target samples with higher consistency are more reliable. The cross-domain classification consistency can be measured by Eq.\ref{entropy}. The lower entropy to classify target samples with a source classifier implies the higher consistency.
\begin{equation}
	H(\boldsymbol{x}^{t};F,C)=-\hspace{-1em}\sum_{i=1}^{|\mathcal{C}_{\mathcal{S}}|+k^\ast}\hspace{-0.8em}C_{i}(F(\boldsymbol{x}^{(t)})) \log C_{i}(F(\boldsymbol{x}^{(t)}))
	\label{entropy}
\end{equation}
%$\forall\boldsymbol{x}^{t}\in\mathcal{T}$, SCDA constructs the subsets in terms of the
where $C_{i}(F(\boldsymbol{x}^{(t)}))$ denotes the softmax value of $\boldsymbol{x}^{t}$ with respect to the $i'$-th class; $k^\ast$=$\mathcal{C}_{\mathcal{D}}$ denotes the optimal estimation of $|\mathcal{C}_{\mathcal{T}}/\mathcal{C}_{\mathcal{S}}|$ in the previous epoch ($k^\ast$=$1$ in the initialization). $\mathcal{C}_{\mathcal{D}}$ indicates the newly discovered target classes and the goal of SCDA is to iteratively update $\mathcal{C}_{\mathcal{D}}$ to approximate $\mathcal{C}_{\mathcal{T}}/\mathcal{C}_{\mathcal{S}}$.

%The goal is to iteratively update $\mathcal{C}_{\mathcal{D}}$ to approximate $\mathcal{C}_{\mathcal{T}}/\mathcal{C}_{\mathcal{S}}$.

Specifically, for each class in $\mathcal{C}_{\mathcal{S}}$, SCDA picks out the samples in with $\mathcal{T}$ the corresponding pseudo label and selects the first half of them with low entropies to construct the target candidate subset of known classes $\hat{\mathcal{T}}_{\rm kn}$. Similarly, as for those in $\mathcal{C}_{\mathcal{D}}$, SCDA also selects the half of them to construct the target candidate subset of implicit classes $\hat{\mathcal{T}}_{\rm im}$.
Obviously, the domain gap between $\hat{\mathcal{T}}_{\rm kn}$ and $\hat{\mathcal{T}}_{\rm im}$ has been erased, and thus, SCDA executes a dynamical cluster algorithm which splits the features of $\hat{\mathcal{T}}_{\rm kn}\cup\hat{\mathcal{T}}_{\rm im}$ by varying the clustering number $k$ then compares their clustering consistency value to obtain an optimal $k$ as the estimation of $|\mathcal{C}_{\mathcal{T}}/\mathcal{C}_{\mathcal{S}}|$.

\textbf{Criteria for the clustering consistency.} We employ two criteria to evaluate the clustering consistency value. The first criterion refers to the elbow method \cite{thorndike1953belongs} widely adopted in clustering analysis. It plots the sum of squared error (SSE) as a function of $k$ to search the elbow point. Specifically, we take the kneedle algorithm \cite{kneedle2011} to locate the point with the ideal cluster number $k^\ast_{\rm elbow}$. The elbow method balances the diversity and the granularity of the clusters, but it can not reflect the prior knowledge of $\mathcal{C}_{\mathcal{S}}$. 

Hence, as a supplement, we compute the clustering accuracy (Eq.\ref{CA}) on $\hat{\mathcal{T}}_{\rm kn}$ to measure the clustering quality. CA measures the clustering accuracy between the clustering assignment and the pseudo label over $\hat{\mathcal{T}}_{\rm kn}$. Higher CA indicates the clustering results are more consistent with $\mathcal{C}_\mathcal{S}$ in the target domain. We select the $k^\ast_{\rm CA}$ with the highest CA.
\begin{equation}
  k_{\rm CA}=\underset{k\in\{1+|\mathcal{C}_{\mathcal{S}}|,\cdots,k_{\max}+|\mathcal{C}_{\mathcal{S}}|\}}{\arg\max}\frac{1}{n}\sum_{i=1}^{n}\mathbf{1}_{\hat{\boldsymbol{y}}_{i}=M(\boldsymbol{c}_{i})}
  \label{CA}
\end{equation}
where $\mathbf{1}$ denotes the indicator function and $M(\boldsymbol{c}_{i})$ is permutation mapping that maps each cluster label $\boldsymbol{c}_{i}$ to the pseudo label $\hat{\boldsymbol{y}}_{i}$ over total $n$ $\boldsymbol{x}_{i}\in\hat{\mathcal{T}}_{\rm kn}$.

SCDA takes $\hat{k}=(k^\ast_{\rm CA}+k^\ast_{\rm elbow})/2$ as the optimal clustering number. Excluding $|\mathcal{C}_{\mathcal{S}}|$ source classes, we consider $k^\ast=\hat{k}-|\mathcal{C}_{\mathcal{S}}|$ as the optimally estimated number of  $\mathcal{C}_\mathcal{T}$/$\mathcal{C}_{\mathcal{S}}$, and use the corresponding clustering assignment to divide $\mathcal{\hat{T}}_{\rm im}$ into $k^\ast$ pseudo classes $\{\hat{\mathcal{T}}_{i}\}^{k^\ast}_{i=1}$. It refers to the update of $\mathcal{C}_{\mathcal{D}}$.
 
\subsubsection{\textbf{Self-supervised open-set adaptation}}\label{step_2_dis}

Step 1 has provided an estimation result for the open-set class discovery, while it is not able to improve OSDA since the result has not been fed back to the domain-invariant feature learning yet. Provided with this, we develop the self-supervised OSDA which enables the classifier $C$ to recognize more target samples that belong to the classes in $\mathcal{C}_{\mathcal{T}}/\mathcal{C}_{\mathcal{S}}$, further improving the OSDA performance. 

\textbf{Restructuring $C$.} Since $\mathcal{C}_{\mathcal{D}}$ dynamically changes to approximate $\mathcal{C}_{\mathcal{T}}/\mathcal{C}_{\mathcal{S}}$, the softmax classifier $C$ is also dynamically restructured in order to classify $\mathcal{C}_{\mathcal{D}}$: Its output dimension alters from $|\mathcal{C}_{\mathcal{S}}|+k^\ast_{t-1}$ to $|\mathcal{C}_{\mathcal{S}}|+k_t^\ast$, in which the first $|\mathcal{C}_{\mathcal{S}}|$ corresponds the number of classes in $\mathcal{C}_\mathcal{S}$ and the latter refers to $k*$ classes in $\mathcal{C}_{\mathcal{D}}$. The parameters are reset and trained in terms of $\mathcal{C}_\mathcal{S}$ and the current $\mathcal{\hat{C}_\mathcal{D}}$.

\textbf{Dynamic class correlation matrix.} The crucial problem is to improve the generalization ability of $F$ and $C$ in terms of the newly discovered classes. To this end, we reconsider the class correlation matrix proposed in Eq.\ref{R}. Indeed, this class decorrelating technique augments various close-set UDAs to reap the transfer gain. However, it is rarely applied in OSDA since its confusion mechanism naturally repels the ``\emph{unknown}'', thus, reducing their confusion would eliminate the intrinsic diversity of $\mathcal{C}_{\mathcal{T}}/\mathcal{C}_{\mathcal{S}}$. This concern found an echo of our motivation, inspiring us to generalize the class correlation matrix to suit OSDA. Specifically, we reconfigure Eq.\ref{R} by using our restructuring softmax classifier output as $\mathbf{\hat{y}}_{i,\cdot}$. Hence, the dimension of $\mathbf{R}$ changes from $(|\mathcal{C}_{\mathcal{S}}|+1)\times(|\mathcal{C}_{\mathcal{S}}|+1)$ to $(|\mathcal{C}_{\mathcal{S}}|+k^\ast)\times(|\mathcal{C}_{\mathcal{S}}|+k^\ast)$, extending the confusion measurement from the known classes in $\mathcal{C}_{\mathcal{S}}$ and the ``\emph{unknown}'' class, to the newly discovered classes in $\mathcal{C}_{\mathcal{D}}$ and the correlation between $\mathcal{C}_{\mathcal{S}}$ and $\mathcal{C}_{\mathcal{D}}$. OSDA with the dynamic class correlation matrix aims to minimize:

\begin{equation}
	\underset{C, F}{\min} \ L_{\rm tcc} = \mathbb{E}_{\{\boldsymbol{x}^{(t)}_i\}^{m}_{i=1}\sim\mathcal{\mathcal{T}}} \ \frac{1}{|\mathcal{C}_{\mathcal{S}}|+{k}^\ast}\sum_{j=1}^{|\mathcal{C}_{\mathcal{S}}|+{k}^\ast}\sum_{{j}'\neq j}^{|\mathcal{C}_{\mathcal{S}}|+{k}^\ast}{\hat{\mathbf{R}}}_{j,{j}'}
	\label{L_tcc}
\end{equation}

Compared with Eq.\ref{L_kcc}, Eq.\ref{L_tcc} iteratively changes its dimension to measure the confusion of $\mathcal{C}_{\mathcal{S}}\cup\mathcal{\hat{C}}_{\mathcal{D}}$. It disambiguates the pseudo class assignment produced by Algorithm.1 and helps $F$ learn more discriminative features for self-supervised open-set adaptation.

To preserve the knowledge from the known classes, we also keep training C with the source samples (Eq.\ref{L_source}). In order to further approach an ideal performance, we simultaneously incorporate the pseudo-labeled target samples drawn from $\hat{{\mathcal{T}}}_{\rm kn}\cup\hat{\mathcal{T}}_{\rm im}$ to learn transferable features:
\begin{equation}
	L_{t} = - \ \mathbb{E}_{(\boldsymbol{x},\boldsymbol{\hat{y}})\sim\hat{{\mathcal{T}}}_{\rm kn}\cup\hat{\mathcal{T}}_{\rm im}} \ \boldsymbol{\hat{y}}^{T}\log \ C(F(\boldsymbol{x}))
	\label{L_target}
\end{equation}
where $\boldsymbol{\hat{y}}$ denotes the corresponding pseudo label. 

\begin{equation}
\mathop{}_{F,C}^{\min} \ L_{\rm s}+L_{\rm t}+L_{\rm tcc}
\label{step_2}
\end{equation}

In summary, the final objective of this step is formulated as Eq.\ref{step_2}. Note that we do not use any hyperparameter to balance each term in all objectives (Eq.\ref{step_1_C}, Eq.\ref{step_1_F} and Eq.\ref{step_2}).

\begin{table*}[ht!]	
  \caption{Results on Office-31 for OSDA. $^{\circ}$ indicates our re-implementation with the officially released code.}
  \footnotesize
  \centering
  \renewcommand\tabcolsep{2.55pt}
  \begin{tabular}{lllllllllllllll}
    \toprule
    \multirow{2}[4]{*}{Method}&\multicolumn{2}{c}{A$\rightarrow $W}&\multicolumn{2}{c}{A$\rightarrow $D}&\multicolumn{2}{c}{D$\rightarrow$W}&\multicolumn{2}{c}{W$\rightarrow $D}&\multicolumn{2}{c}{D$\rightarrow $A}&\multicolumn{2}{c}{ W$\rightarrow $A }&\multicolumn{2}{c}{Avg} \\
    \cmidrule{2-15}&\multicolumn{1}{c}{OS} &\multicolumn{1}{c}{OS*}&\multicolumn{1}{c}{OS} &\multicolumn{1}{c}{OS*}&\multicolumn{1}{c}{OS} &\multicolumn{1}{c}{OS*}&\multicolumn{1}{c}{OS} &\multicolumn{1}{c}{OS*}&\multicolumn{1}{c}{OS} &\multicolumn{1}{c}{OS*}&\multicolumn{1}{c}{OS} &\multicolumn{1}{c}{OS*}&\multicolumn{1}{c}{OS} &\multicolumn{1}{c}{OS*} \\
    \midrule
    OSBP &86.5$\pm$2.0&87.6$\pm$2.1&88.6$\pm$1.4&89.2$\pm$1.3&97.0$\pm$1.0&96.5$\pm$0.4&97.9$\pm$0.9&98.7$\pm$0.6&88.9$\pm$2.5&90.6$\pm$2.3&85.8$\pm$2.5&84.9$\pm$1.3&90.8 &91.3 \\
    STA  &89.5$\pm$0.6&92.1$\pm$0.5&93.7$\pm$1.5&96.1$\pm$0.4&97.5$\pm$0.2&96.5$\pm$0.5&99.5$\pm$0.2&99.6$\pm$0.1&89.1$\pm$0.5&93.5$\pm$0.8&87.9$\pm$0.9&87.4$\pm$0.6&92.9 &94.1 \\
    TIM  &91.3$\pm$0.7&93.2$\pm$1.2&94.2$\pm$1.1&\textbf{97.1$\pm$0.8}&96.5$\pm$0.5&97.4$\pm$0.7&99.5$\pm$0.2&99.4$\pm$0.3&90.1$\pm$0.2&91.5$\pm$0.2&88.7$\pm$1.3&88.1$\pm$0.9&93.4 &94.5 \\
    JPOT  & 92.8$\pm$0.6&92.2$\pm$0.4&95.2$\pm$0.9&96.0$\pm$0.6&98.1$\pm$0.3&96.2$\pm$0.4&99.5$\pm$0.1&98.6$\pm$0.2&\textbf{93.0$\pm$0.7}&\textbf{94.1$\pm$0.4}&88.9$\pm$1.0&88.4$\pm$0.4&94.6 &94.3 \\

SHOT$^{\circ}$ & 88.8$\pm$0.7& 91.4$\pm$0.4& 90.3$\pm$0.5& 92.6$\pm$0.3& 96.2$\pm$0.4& 97.0$\pm$0.4& 97.4 $\pm$0.3&97.9$\pm$0.5& 91.6$\pm$0.5& 93.4$\pm$0.7& 91.2$\pm$0.5& 93.5$\pm$0.4& 92.6 &94.3 \\

PGL$^{\circ}$ & 89.2$\pm$0.5& 90.1$\pm$0.7& 89.6$\pm$0.8& 91.6$\pm$0.5& 95.3$\pm$0.4& 95.1$\pm$0.5& 96.7$\pm$0.5& 97.6$\pm$0.6& 71.0$\pm$0.9& 72.0$\pm$0.5& 73.0$\pm$0.4& 77.6$\pm$0.6& 85.8& 87.4\\
  
    \textbf{SCDA}&\textbf{95.7$\pm$0.1}&\textbf{97.5$\pm$0.2}&\textbf{95.9$\pm$0.4}&96.5$\pm$0.4&\textbf{99.2$\pm$0.1}&\textbf{99.7$\pm$0.1}&\textbf{99.8$\pm$0.1}&\textbf{100$\pm$0.0}&92.1$\pm$0.1&93.7$\pm$0.3&\textbf{92.2$\pm$0.1}&\textbf{93.6$\pm$0.1}&\textbf{95.8}&\textbf{96.8} \\
    \bottomrule
  \end{tabular}
  \label{tab_office31}
\end{table*}

\begin{table*}[ht!]
  \caption{Results on Office-Home for OSDA. $^{\triangle}$ indicates the method does not report the variance of their results.}
  \footnotesize
  \centering
  \renewcommand\tabcolsep{3.2pt}
  \begin{tabular}{lcccccccccccccc}
    \toprule
    Method & Ar$\rightarrow$Cl & Pr$\rightarrow$Cl & Rw$\rightarrow$Cl & Ar$\rightarrow$Pr & Cl$\rightarrow$Pr & Rw$\rightarrow$Pr & Cl$\rightarrow$Ar & Pr$\rightarrow$Ar & Rw$\rightarrow$Ar & Ar$\rightarrow$Rw & Cl$\rightarrow$Rw & Pr$\rightarrow$Rw & \multicolumn{1}{l}{Avg} \\
    \midrule
    OSBP  & 56.7$\pm$1.9 & 51.5$\pm$2.1 & 49.2$\pm$2.4 & 67.5$\pm$1.5 & 65.5$\pm$1.5 & 74.0$\pm$1.5 & 62.5$\pm$2.0 & 64.8$\pm$1.1 & 69.3$\pm$1.1 & 80.6$\pm$0.9 & 74.7$\pm$2.2 & 71.5$\pm$1.9 & 65.7 \\
    STA   & 58.1$\pm$0.6 & 53.1$\pm$0.9 & 54.4$\pm$1.0 & 71.6$\pm$1.2 & 69.3$\pm$1.0 & 81.9$\pm$0.5 & 63.4$\pm$0.5 & 65.2$\pm$0.8 & 74.9$\pm$1.0 & 85.0$\pm$0.2 & 75.8$\pm$0.4 & 80.8$\pm$0.3 & 69.5 \\
    TIM   & 60.1$\pm$0.7 & 54.2$\pm$1.0 & 56.2$\pm$1.7 & 70.9$\pm$1.4 & 70.0$\pm$1.7 & 78.6$\pm$0.6 & 64.0$\pm$0.6 & 66.1$\pm$1.3 & 74.9$\pm$0.9 & 83.2$\pm$0.9 & 75.7$\pm$1.3 & 81.3$\pm$1.4 & 69.6 \\
    JPOT & 59.6$\pm$0.5 & 54.2$\pm$0.7 & 54.6$\pm$0.9 & 72.3$\pm$1.1 & 70.1$\pm$0.6 & 82.1$\pm$0.9 & 62.9$\pm$0.7 & 68.3$\pm$0.8 & 75.1$\pm$1.1 & 84.8$\pm$0.4 & 77.4$\pm$0.5 & 81.2$\pm$0.4 & 70.2\\
    SHOT$^{\triangle}$  & \multicolumn{1}{l}{64.5$\pm$0.0} & \multicolumn{1}{l}{\textbf{59.3$\pm$0.0}} & \multicolumn{1}{l}{64.6$\pm$0.0} & \multicolumn{1}{l}{\textbf{80.4$\pm$0.0}} & \multicolumn{1}{l}{\textbf{75.4$\pm$0.0}} & \multicolumn{1}{l}{82.3$\pm$0.0} & \multicolumn{1}{l}{63.1$\pm$0.0} & \multicolumn{1}{l}{65.3$\pm$0.0} & \multicolumn{1}{l}{69.6$\pm$0.0} &  \multicolumn{1}{l}{84.7$\pm$0.0} & \multicolumn{1}{l}{81.2$\pm$0.0} & \multicolumn{1}{l}{83.3$\pm$0.0} & 72.8  \\
    PGL$^{\triangle}$  & \multicolumn{1}{l}{\textbf{61.6$\pm$0.0}}& \multicolumn{1}{l}{58.4$\pm$0.0}& \multicolumn{1}{l}{\textbf{65.0$\pm$0.0}}& \multicolumn{1}{l}{77.1$\pm$0.0}& \multicolumn{1}{l}{72.0$\pm$0.0} & \multicolumn{1}{l}{83.0$\pm$0.0}& \multicolumn{1}{l}{68.8$\pm$0.0}& \multicolumn{1}{l}{72.2$\pm$0.0} & \multicolumn{1}{l}{\textbf{78.6$\pm$0.0}}& \multicolumn{1}{l}{\textbf{85.9$\pm$0.0}}&  \multicolumn{1}{l}{82.8$\pm$0.0}& \multicolumn{1}{l}{82.6$\pm$0.0}&   74.0        \\
    \textbf{SCDA} & 59.9$\pm$0.3 & 59.0$\pm$0.3 & 62.8$\pm$0.5 & 79.6$\pm$0.4 & 73.8$\pm$1.0 & \textbf{83.7$\pm$0.8} & \textbf{70.9$\pm$0.5} & \textbf{72.3$\pm$0.6} & 75.5$\pm$0.4 & 85.3$\pm$0.6 & \textbf{82.9$\pm$0.3} & \textbf{85.7$\pm$0.9} & \textbf{74.3} \\
    \bottomrule
  \end{tabular}
  \label{tab_officehome}
\end{table*}

\section{Experiment} 

In this section, we evaluate SCDA on three benchmarks to demonstrate its superior performance on both OSDA and discovering implicit classes.

\textbf{Benchmarks.} We use two famous datasets: \textbf{Office-31} and \textbf{Office-Home}, and choose the same label sets of classes to build $\mathcal{C}_{\mathcal{S}}$ and $\mathcal{C}_{\mathcal{T}}$ following \cite{liu2019separate}. 
Besides, we introduce the challenging \textbf{DomainNet}. To simulate a real-world adaptation scenario, we combine the \emph{Real} and \emph{Clipart} domains in DomainNet with the \textbf{Rw} and \textbf{Cl} in Office-Home, respectively, to build two target domains \textbf{Rw$^{\star}$} and \textbf{Cl$^{\star}$}. After merging the same categories, the combined target domains have 362 categories, 279 classes of which are \textbf{DomainNet}-specific. Then we randomly select $1/4$ classes from them to induce the scarcity: we select 10 samples for each of the classes and abandon the rest. It breeds a benchmark \textbf{DomainNet$^{\star}$} with extremely imbalanced target domains \textbf{Rw$^{\star}$} and \textbf{Cl$^{\star}$}.

\textbf{Baselines.} We compare SCDA with a variety of state-of-the-art OSDA approaches, including \textbf{OSBP}, \textbf{STA}, \textbf{TIM}, \textbf{JOPT}, \textbf{SHOT}, and \textbf{PGL}. We are also interested in the performance of discovering implicit classes. To this, we compare SCDA's class-discovering ability with some state-of-the-art baselines, \emph{i.e.}, Silhouette coefficient (\textbf{SC}) and \textbf{DTC}.

\textbf{Evaluation Criteria.}
For a fair comparison, we employ two evaluation metrics in line with \cite{Saito2018Open}, \emph{i.e.}, \textbf{OS: }averaging the class-wise target accuracy for all the classes including the unknown as one class; \textbf{OS*: }averaging the class-wise target accuracy only on known classes. Besides, in terms of class discovery, we compare SCDA with SC and DTC to estimate the number of unknown implicit classes $k^\ast$. 

\textbf{Implementation.}  Following \cite{Saito2018Open}, we evaluate all methods on all datasets with ResNet-50 pre-trained on ImageNet as the backbone. We implement SCDA in PyTorch and use momentum SGD with a learning rate of $10^{-3}$. More implementation details can be found in supplementary material \ref{SM_experimental_details}.

\subsection{Results for OSDA}

\noindent \textbf{\textbf{Office-31} and \textbf{Office-Home.}} 
In Table \ref{tab_office31}, SCDA outperforms other baselines on most transfer tasks in \textbf{Office-31} with significant margins. In the hard tasks, \emph{e.g.}, A$\rightarrow$W, SCDA outperforms the second with a larger gap (\textbf{2.9\%}).
As illustrated in Table \ref{tab_officehome}, our SCDA still achieves the best performance in \textbf{Office-Home} dataset. 
Notably, the second best model in \textbf{Office-31} (TIM) and \textbf{Office-Home} (PGL) both perform poorly in the other dataset. It is probably due to the changing setup of \emph{unknown}. In a comparison, SCDA is designed to analyze the inter-class structure of $\mathcal{C}_\mathcal{T}$/$\mathcal{C}_\mathcal{S}$. Thus, our method presents the more robust generalization ability in OSDA.

\begin{table}
  \centering                    
  \setlength\tabcolsep{3.0pt}
  \footnotesize 
  \caption{OSDA from \textbf{OfficeHome} to \textbf{DomainNet$^{\star}$}}
  \begin{tabular}{ccccccccccc}
    \toprule
    \multirow{2}[3]{*}{Method} & \multicolumn{2}{c}{Ar$\rightarrow$Rw$^{\star}$} & \multicolumn{2}{c}{Ar$\rightarrow$Cl$^{\star}$} & 	\multicolumn{2}{c}{Pr$\rightarrow$Rw$^{\star}$} & 	\multicolumn{2}{c}{Pr$\rightarrow$Cl$^{\star}$} & \multicolumn{2}{c}{Avg} \\
    \cmidrule{2-11}   & OS   & OS* & OS   & OS*   & OS    &  OS*   & OS   & OS*  & OS   & OS*  \\
    \midrule
    OSBP  &  58.1 & 57.8 &33.0 & 32.3 &  59.5  &  59.3 & 30.5 & 29.8 & 45.4 & 44.8 \\
    STA   &  60.5 &  60.4 & 40.1 & 39.6 & 59.1 & 59.0 & 32.3 &33.9 & 48.0 & 48.2 \\
    SHOT &64.6 & 65.1 & \textbf{45.2} & \textbf{45.7} & 65.4& 65.9 & 40.3& 40.4 & 53.9 & 54.3 \\
    Ours  &  \textbf{67.8} &  \textbf{68.0} & 44.2  &  44.3  & \textbf{68.2} & \textbf{69.1} & \textbf{40.9} & \textbf{41.0} & \textbf{55.3} & \textbf{55.6} \\
    \bottomrule
  \end{tabular}
  \label{wilddata}%
\end{table}

\begin{table}
  \centering
  \caption{Unknown categories number estimation results}
  \renewcommand\tabcolsep{3.0pt}
  \footnotesize
  \begin{tabular}{lccccccc}
    \toprule
    \multirow{2}[4]{*}{Dataset} & \multirow{2}[4]{*}{GT} & \multicolumn{2}{c}{SC} &  \multicolumn{2}{c}{DTC} & \multicolumn{2}{c}{Ours} \\
    \cmidrule{3-8}   &  &  $\hat{k}$     &  Error & $\hat{k}$     &  Error &  $\hat{k}$     &  Error \\
    \midrule
    
    Office-31 & 11    & 8$\sim$33 & 6.8 & 4$\sim$9 & 4.2   & 9$\sim$11  & \textbf{0.9} \\
    Office-Home & 40   & 0$\sim$7 & 38.5 & 9$\sim$23 & 21.4  & 32$\sim$37 & \textbf{6.1} \\
    DomainNet$^{\star}$  & 297     & 5$\sim$8   & 290.7  & 46$\sim$71   & 238.2   & 227$\sim$255   & \textbf{58.5} \\
    \bottomrule
  \end{tabular}
  \label{tab_e_k}	
\end{table}

\noindent \textbf{Real-world Scenarios.} To further investigate the baselines in more complicated real world applications, we set up the transfer tasks from \textbf{Pr} and \textbf{Ar} in \textbf{Office-Home} to the challenging blending-target domains \textbf{Rw$^{\star}$} and \textbf{Cl$^{\star}$} in \textbf{DomainNet$^{\star}$}. As shown in Table \ref{wilddata}, although \textbf{Rw$^{\star}$} and \textbf{Cl$^{\star}$} are noisy, and extremely imbalanced, SCDA still achieves the state of the art in $3$ from $4$ transfer combination. Besides, SCDA also presents a faster convergence rate and a higher upper-bound performance in the complicated scenarios (see Fig.\ref{pic_iter} in SM).

\subsection{Results for Implicit Class Discovery}

In Table \ref{tab_e_k}, we report the results for unknown class number estimation by SC, DTC, and SCDA. SC performs the worst across all the benchmarks. DTC is poor in \textbf{Office-Home} and \textbf{DomainNet$^{\star}$} with numerous implicit classes. By contrast, SCDA shows surprisingly accurate results to estimate the class number in \textbf{Office-31}, where the average error is less than 1. Despite the large implicit class number in \textbf{Office-Home} and \textbf{DomainNet$^{\star}$}, SCDA produces a low average error. The results validate the reliability of SCDA to estimate the implicit class number. More experiments for class discovery can be found in supplementary material \ref{class_discovery}.

\begin{figure}
	\includegraphics[width=0.8 \columnwidth]{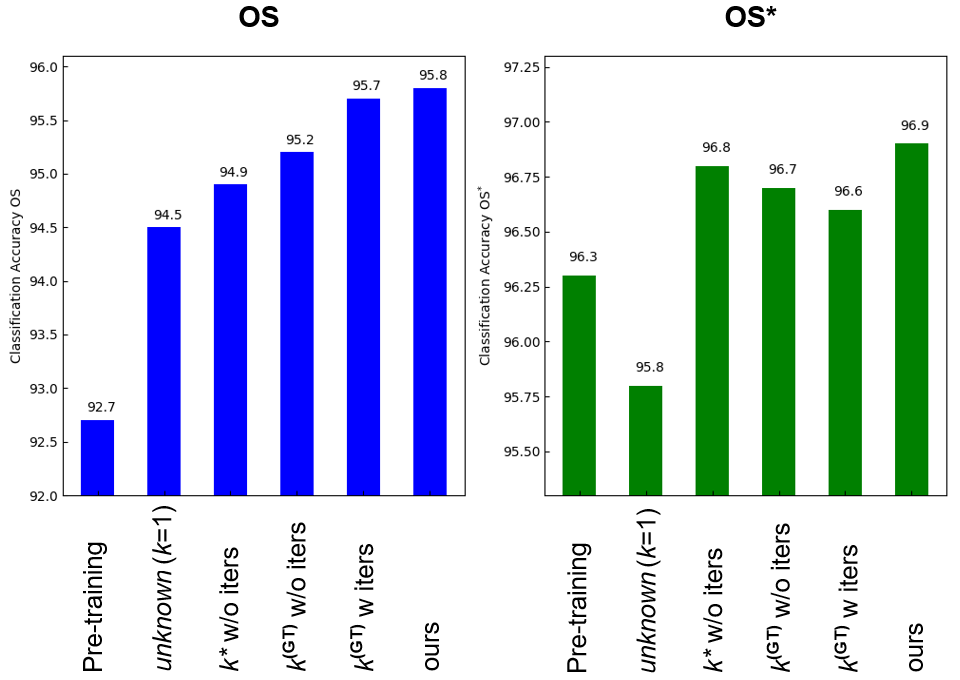}
	\centering
	\caption{ Average results of SCDA}
	\label{pic_ablation}
\end{figure}

\subsection{Ablation study}

The motivation of SCDA rises from the conjecture that the unobserved class discovery may help OSDA. Our ablation is designed to justify the conjecture. In particular, we compare SCDA with the following modifications. (\textbf{1}) \textbf{Pre-training}: we use the network only pre-trained by subsection \ref{step_0}. (\textbf{2}) \textbf{\emph{unknown} ($k$=1)}: we utilize the pseudo labels to update the model. However, without class discovery, we regard them as a single negative class \emph{unknown}. (\textbf{3}) \textbf{$k^{*}$ w/o iters}: SCDA trains the model with pseudo discovered classes but without further iteration. (\textbf{4}) \textbf{$k^{(GT)}$ w/o iters}: we provide with the true number of $\mathcal{C}_\mathcal{T}$ but without further iteration. (\textbf{5}) \textbf{$k^{(GT)}$ w iters}: the algorithm is provided with $k^{(GT)}$, then we alternatively train the model.
	
As illustrated in Figure \ref{pic_ablation}, we report the average OS and OS* across all six transfer tasks in Office-31. 
By comparing (\textbf{2}) with (\textbf{3}), discovering the unobserved classes has a better performance than regarding them as the ``\emph{unknown}''. It further verifies our motivation: discovering the structure of the unobserved classes can improve the performance of OSDA.
The results of (\textbf{3}-\textbf{5}) draw an interesting conclusion. Without further iterations, training with $k^{(GT)}$ outperforms training with the estimated class number. However, their results are almost the same when we train model iteratively. It suggests that the precise prediction of the unobserved classes number and the iterative optimization are both important and their combination play a key role in addressing OSDA. Besides, more \textbf{analysis} of the ratio of unobserved classes and \textbf{visualization} are illustrated in Supplementary Material \ref{analysis}.

\section{Conclusion}
In this paper, we pay attention to a nontrivial challenge in OSDA: discovering all implicit classes in the unknown target samples. The mixed unknown chunk conceives category mismatching risk. To tackle the problem, we propose Self-supervised Class-Discovering Adapter (\textbf{SCDA}). SCDA utilizes adversarial learning to preliminarily separate unknown target samples. Then, SCDA employs an alternate approach to discover novel target categories and update our model with the discovery results. Through extensive empirical evaluations, we demonstrate the superiority of our SCDA by the state-of-the-art OSDA performance and the remarkable ability to discover unknown implicit classes.

\bibliographystyle{IEEEbib}
\bibliography{SCDA_arxiv}

\clearpage

\begin{figure*}[t]
  \centering
  \huge{\textbf{Supplementary Material of Open Set Domain Adaptation \\ By Novel Class Discovery}}
\end{figure*}

\section{The pipeline of SCDA}
The pipeline of SCDA is found in Algorithm.\ref{Training}

\setcounter{algorithm}{1}
\begin{algorithm}[t]			
	\caption{Self-supervised Class-Discovering Adapter} \label{Training} 
				
	{\bf Input:} Labeled source dataset $\mathcal{S}$; Unlabeled target dataset $\mathcal{T}$; initiated feature extractor $F$ and classifier $C$; max iteration epoch $E$.\\
	{\bf Output:} Well-trained $F^{*}$ and $C^{*}$; The predicted number of unknown classes $k^\ast$.
				
	\begin{algorithmic}[1]	
		\STATE \textbf{Pre-train} $F$ by Eq.5 \textbf{and} $C$ by Eq.6
		\STATE \textbf{for} $0:E$ \textbf{do}
		\STATE \qquad \textbf{Step 1: Implicit Class Discovery}			\STATE \qquad Get $k^\ast$, $\hat{\mathcal{T}}_{\rm sh}$, $\hat{\mathcal{T}}_{\rm im}$ and $\{\mathcal{\hat{T}}_i\}^{k^\ast}_{i=1}$ by Algorithm.1.
		\STATE \qquad \textbf{Step 2: Self-supervised OSDA}
		\STATE \qquad Reset $C$ with the dimension of $|\mathcal{C}_{\mathcal{S}}|+{k}^\ast$. 
		\STATE \qquad Retrain $F$ and $C$ by Eq.11.
		\STATE \textbf{End for}
		\STATE \textbf{Return} $\{\mathcal{\hat{T}}_i\}^{k^\ast}_{i=1}$; ${k}^\ast$; $F^{*}=F$; $C^{*}=C$.
					
	\end{algorithmic}
\end{algorithm}

\section{Experimental Implementation}\label{SM_experimental_details}
In this section, we present more experimental details. The code will be released later.

\subsection{Architectures}

\begin{figure}[htbp]
		\centering
		\includegraphics[width=1\columnwidth]{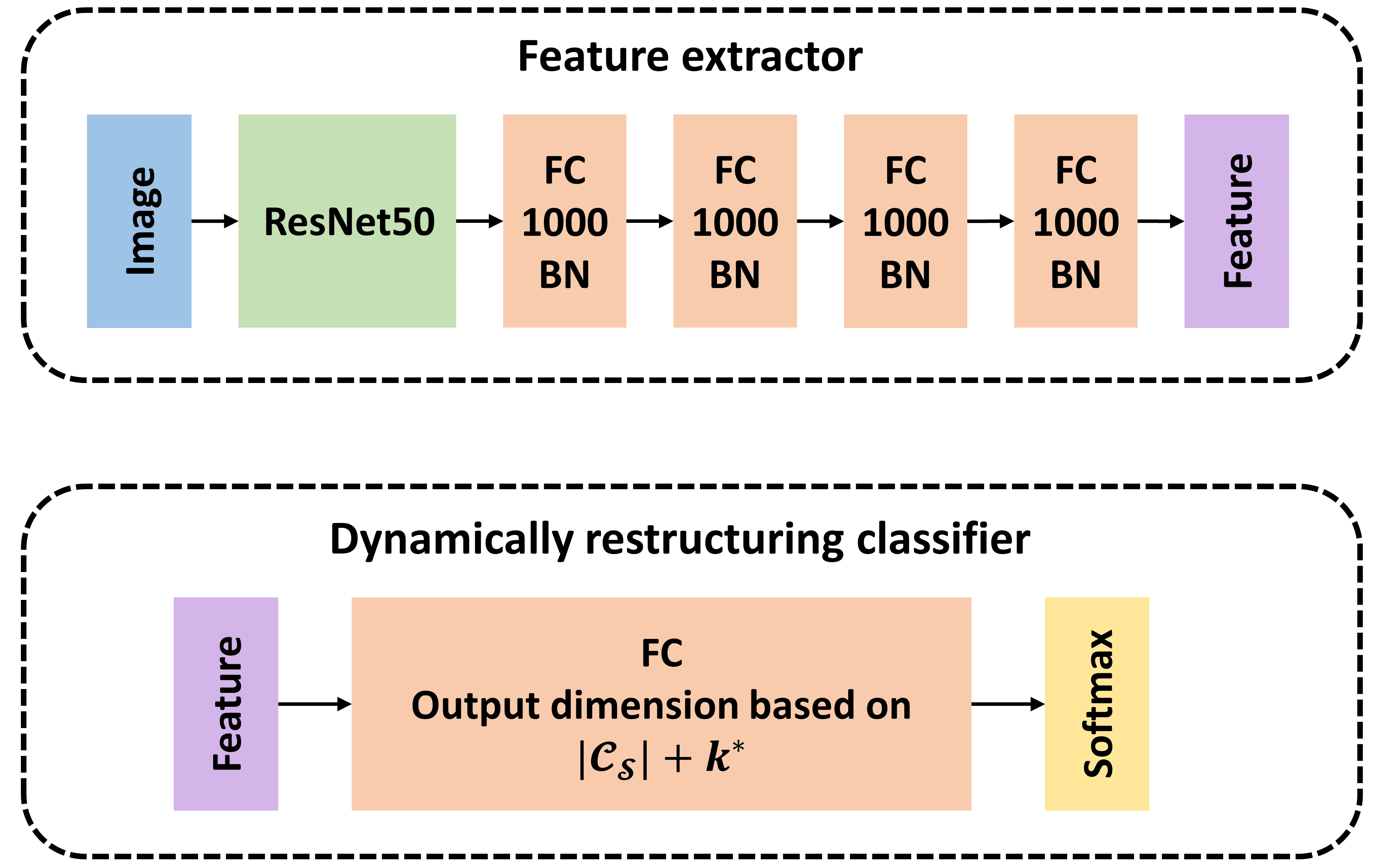}
		\caption{The feature extractor $F$ and dynamically restructuring classifier $C$ we used in the experiments about visual recognition. (Best viewed in color)}
		\label{pic_architecture}
	\end{figure}

For fair comparisons with existing methods \cite{Saito2018Open, liu2019separate, kundu2020towards}, we evaluate SCDA and all baselines on Office-31 dataset \cite{saenko2010adapting}, Office-Home dataset \cite{venkateswara2017deep}, and DomainNet* \cite{peng2019moment} with ResNet-50 as the backbone. ResNet-50 is pre-trained on ImageNet \cite{russakovsky2015imagenet}, then replaced the last fc layer of ResNet-50 with four fully connected layers with batch normalization \cite{ioffe2015batch}. The detail architectures of $F$ and $C$ are illustrated in Figure \ref{pic_architecture}.

\subsection{Network Training}
We implement the framework in PyTorch and use momentum SGD with learning rate of $10^{-3}$, the momentum is set as 0.9 and the weight decay as $5\times 10^{-4}$. During the Self-supervised OSDA pre-training step (\emph{i.e.}, Eq.6,7 in the paper), we do not update the parameters of the backbone network. When turning to the alternative steps, the parameters of the feature extractor ($F$) would be iteratively updated while for the restructuring classifier ($C$), its parameter would be reset and trained from scratch for each iteration in order to suit its dynamical structure. The hyper-parameters are shown in Table \ref{hyper}.

\subsection{Acceleration.} It is worth noticing that SCDA is an iterative algorithm where each iteration relies on Algorithm.1, which estimates the implicit class number by iteratively performing $k$-means. Given a linear computation complexity of a $k$-means implementation in Pytorch, the complexity of Algorithm.1 refers to 
\begin{equation*}
O(Td_{F}k^2_{max}n_{\rm ct})
\end{equation*}where $T$ denotes the max iteration number of the $k$-means implementation; $d_{F}$ denotes the feature dimension for clustering; $n_{\rm ct}$ denotes the total number of the confidential target subsets $\hat{\mathcal{T}}_{\rm ex}\cup\hat{\mathcal{T}}_{\rm im}$; $k_{max}$ indicates the maximum estimation to the number of all implicit classes. As the target domain $\mathcal{T}$ become diverse, $k_{max}$ will increase and slow the algorithm. To this end, PCA is applied to reduce the output feature dimension $d_{F}$, and we also employ GPU-driven $k$-means++ \cite{arthur2007k} to seed the cluster centroids and run $k$-means, which implies a trivial $T$ to achieve the convergence in practice. Beyond these operations, to accelerate the whole SCDA algorithm, we also encourage the practitioners to reduce the max number of outer loop iteration $E$ in terms of a benchmark with diverse implicit classes, \emph{e.g.}, DomainNet.

As shown in Table.\ref{hyper}, SCDA allows a large $k_{max}$ setup to search the implicit classes as many as possible. Specifically, the main learning procedure of SCDA consumes 3 hours for OfficeHome and 12 hours for DomainNet$^\ast$ by using a single Geforce GTX TITAN-X.

\begin{table*}[htbp]
	\centering
	\caption{Implicit categories number estimation results of DomainNet* and Office-31}
	\setlength\tabcolsep{3.0pt} 
	\begin{tabular}{lccccccccccccc}
	\toprule
	
	\multicolumn{1}{l|}{Dataset} & \multicolumn{5}{c|}{DomainNet*} & \multicolumn{8}{c}{Office-31} \\
	\midrule
	\multicolumn{1}{l|}{Task} & Ar$\rightarrow$Rw* & Ar$\rightarrow$Cl*& Pr$\rightarrow$Rw*& Pr$\rightarrow$Cl*& \multicolumn{1}{c|}{Error} & A$\rightarrow$W  & A$\rightarrow$D  & D$\rightarrow$W  & W$\rightarrow$D  & D$\rightarrow$A  &  W$\rightarrow$A  & Avg & Error \\
	\midrule
	SC   & 8.3  & 5.0 &  6.7 &5.0 & 290.7 & 6.3 & 21.3  & 36.7  & 18   & 13.3  & 11.0   & 17.8 & 6.8  \\
	DTC & 71.0 & 55.7 & 62.3 & 46.0  & 238.2 & 8.0     & 6.3   & 5.3   & 4.7   & 8.3   & 8.3   & 6.8 & 4.2 \\
	SCDA & 227.3 & 237.3 &  255.3 & 234.0 &58.5 & 10.7  & 11.0    & 9.7   & 10.7  & 9.3   & 9.3   & 10.1  & 0.9 \\
	GT    & 6 & 6& 6& 6   & - & 11    & 11    & 11    & 11    & 11    & 11    & 11   & - \\
	\bottomrule
	\end{tabular}%
	\label{tab_1}%
\end{table*}%

\begin{table*}[htbp]
	\centering
	\caption{Implicit categories number estimation results of Office-Home}
	\setlength\tabcolsep{4.3pt} 
	\begin{tabular}{lcccccccccccccc}
	\toprule
	Task  & A$\rightarrow$C  & P$\rightarrow$C  & R$\rightarrow$C  & A$\rightarrow$P  & C$\rightarrow$P  & R$\rightarrow$P  & C$\rightarrow$A  & P$\rightarrow$A  & R$\rightarrow$A  & A$\rightarrow$R  & C$\rightarrow$R  & P$\rightarrow$R  & Avg &Error  \\
	\midrule
	SC   & 1.3    & 0.3  & 0.3  & 0.3  & 6.3  & 5.3    & 1.3  & 0.0    & 0.0  &  0.0 &   2.3   &  0.7  & 1.5 & 38.5\\
	DTC   & 20.0    & 13.3  & 22.3  & 19.3  & 22.3  & 17.0    & 18.7  & 18.0    & 16.7  & 17.7  & 21.0    & 17.0    & 18.6 & 21.4 \\
	SCDA & 33.3  & 34.3  & 35.3  & 34.7  & 32.7  & 36.3  & 32.3  & 31.3  & 31.7  & 33.3  & 36.3  & 35.3  & 33.9 & 6.1 \\
	GT    & 40    & 40    & 40    & 40    & 40    & 40    & 40    & 40    & 40    & 40    & 40    & 40    & 40  & - \\
	\bottomrule
	\end{tabular}%
	\label{tab_2}%
\end{table*}%

\begin{table}
  \setlength\tabcolsep{3.0pt} 
  \centering
  \caption{The hyper-parameters setting in our experiment.}
  \begin{tabular}{lccc}
    \toprule
    \small 
    Dataset & \footnotesize Office-31 & \footnotesize Office-Home & \footnotesize DomainNet*\\
    \midrule
    \small backbone  & \small ResNet-50 & \small ResNet-50 & \small ResNet-50 \\
    batch size     & 32    & 32 & 64 \\
    lr  & 0.001 & 0.001 & 0.001 \\
    image size  &  227$\times $227   &   227$\times $227 &  227$\times $227 \\
    $E$    & 400   & 400 & 40 \\
    $k_{max}$     & 40    & 120 & 650 \\
    \bottomrule
  \end{tabular}%
  \label{hyper}%
\end{table}%

\subsection{Evaluation Criteria.}
For a fair comparison, we employ two OSDA evaluation metrics in line with \cite{Saito2018Open, liu2019separate}, i.e., \textbf{OS: }averaging the class-wise target accuracy for all the classes including the unknown as one class; \textbf{OS*: }averaging the class-wise target accuracy only on shared classes. 

In terms of implicit class discovery, we evaluate our algorithm according to two criterion: 
\begin{enumerate}
	\item \textbf{Implicit class number ($k^\ast$) estimation}: How many the implicit classes the algorithm has estimated?
	\item \hspace{-0.2pt}\textbf{Implicit class correspondence}: \hspace{-0.2pt}How \hspace{-0.2pt}many the real implicit classes the algorithm-discovered classes have corresponded to?
\end{enumerate}

Towards the first criteria, we follow \cite{han2019learning} and run the evaluated algorithms three times for each transfer task, then report their mean of the estimated implicit class number, \emph{i.e.}, $k^\ast$. To this, Avg indicates the average result of all the transfer tasks in each dataset. We compare it with the ground truth number of the implicit classes (GT) and their absolute difference between Avg and GT (smaller indicates better estimation performance, see Table.\ref{tab_1} ,\ref{tab_2}).

Towards the second criteria, we compute the probabilities of the unknown samples and categorized them into the discovered implicit classes in terms of the specific algorithm. For the samples chosen in each discovered class, we sort them by the prediction confidence then choose the top $n$ samples with the highest probabilities. So provided overall $nk^\ast$ samples, we check how many real implicit classes they have referred to. Obviously, the value is less than the ground truth (GT) number of real implicit classes. The value closer to the ground truth indicates better (see Table.\ref{tab_class}).

\begin{table}[!htp]
	\setlength\tabcolsep{4.0pt} 
	\centering
	\caption{Matching real implicit classes in terms of top-$n$ evaluation}
	
	\begin{tabular}{lcccccccc}
	\toprule
	\multirow{2}[4]{*}{Dataset} & \multicolumn{3}{c}{DTC} & \multicolumn{3}{c}{SCDA} & GT\\
	
	\cmidrule{2-8}  & $n$=1   & $n$=3   & $n$=5  & $n$=1   & $n$=3   & $n$=5 & GT\\
	\midrule
	Office-31 & 4& 7.6& 8.9& 8.8   & 9.3   & 10 & 11 \\
	Office-Home & 10.4 &20.2 & 25 & 20.6  & 26.7  & 30 & 40 \\
	DomainNet* & 42.4 & 91.4 & 121.5 &  117    &  141    & 153.5 & 297 \\
	\bottomrule
	\end{tabular}%
	\label{tab_class}%
	\end{table}%

\section{Results}
\begin{figure}
	\centering
	\includegraphics[width=1 \columnwidth]{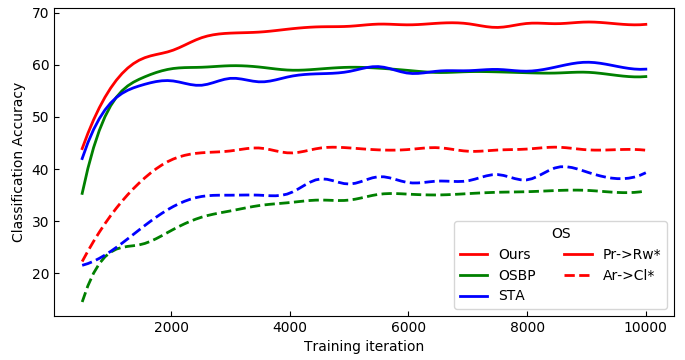}
	\caption{The OS in task Pr$\rightarrow$Rw* and Ar$\rightarrow$Cl* during training}
 %调整图片与上文的垂直距离
	\label{pic_iter}
\end{figure}

In this section, we have provided more evaluation results skipped in the paper. The faster convergence and a higher upper-bound performance in the complicated scenarios are shown in Figure.\ref{pic_iter}.

\subsection{Results for implicit class discovery.}\label{class_discovery}
Here we detailed the evaluation results of implicit class discovery in three OSDA benchmarks Office-31, Office-Home and DomainNet*.

\textbf{Implict class number ($k^\ast$) estimation.} SC \cite{kaufman2009finding} and DTC \cite{han2019learning} are introduced as the baselines besides of our algorithm. SC (Silhouette Coefficients) is a method of validation of consistency within clusters of data and Kaufman et al. \cite{kaufman2009finding} introduced it to estimate the clusters $k$. Deep Transfer Clustering (DTC) \cite{han2019learning} clusters unlabeled data by DEC \cite{xie2016unsupervised} with a pretrained model and transfers knowledge from the set of source-known classes to implicit target samples.

In Table \ref{tab_1} and Table \ref{tab_2}, we report the results of the estimated $k^{*}$ of all transfer tasks in three datasets. As shown in Table \ref{tab_1} and Table \ref{tab_2}, SC performed the worst across all benchmarks and totally failed in Office-Home. DTC poorly performed in Office-Home and DomainNet*. With the number of implicit classes increasing, the results of SC and DTC both became worse. Compared with DTC and SC, our SCDA show the minimum error and the most stable performance. It validates the effectiveness of our approach.

\textbf{Implicit classes corrspondence.}
We further investigate how many implicit classes SCDA may discover. To be specific, after training, SCDA gets a well-trained classifier $C^{*}$ with output dimension $|\mathcal{C}_{s}|+k^{*}$(\emph{e.g.}, $m+k^{*}$), where $k^{*}$ indicates $k^{*}$ novel classes. For each novel class, we consider the probabilities of the unknown samples belonging to this class according to the softmax output of the restructuring classifier. We take them to feed the implicit class correspondence in the second criteria. As for DTC, we choose the top $n$ samples with the highest probability of assigning data points to a class-aware cluster, and check how many implicit classes they have corresponded. We report the average results of all transfer tasks in each dataset.

As shown in Table \ref{tab_class}, with the identical $n$, our method has corresponded to more implicit classes. Especially when $n$=1, for all datasets, our SCDA is twice the number of the discovered classes than DTC.

\begin{figure}

	\centering
	\includegraphics[width=0.8 \columnwidth]{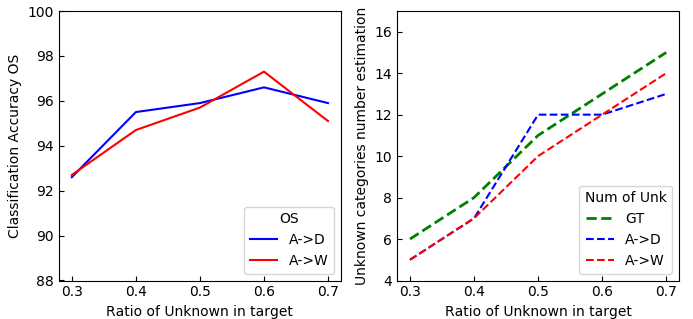}
	\caption{Classification and class discovery performance of \textbf{SCDA} in A$\rightarrow$W and A$\rightarrow$D when the ratio of unknown classes varies.}
	\label{pic_ratio1}
	\centering
	\includegraphics[width=0.8 \columnwidth]{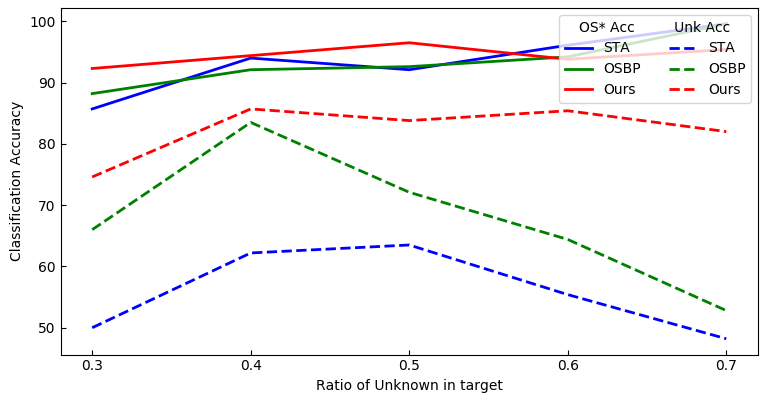}
	
	\caption{OS* and unknown samples accuracy in A$\rightarrow$W when the ratio of unknown classes changes.}
		%调整图片与上文的垂直距离
	\label{pic_ratio2}

\end{figure}

\begin{figure}
	\centering
	\includegraphics[width=0.9\columnwidth]{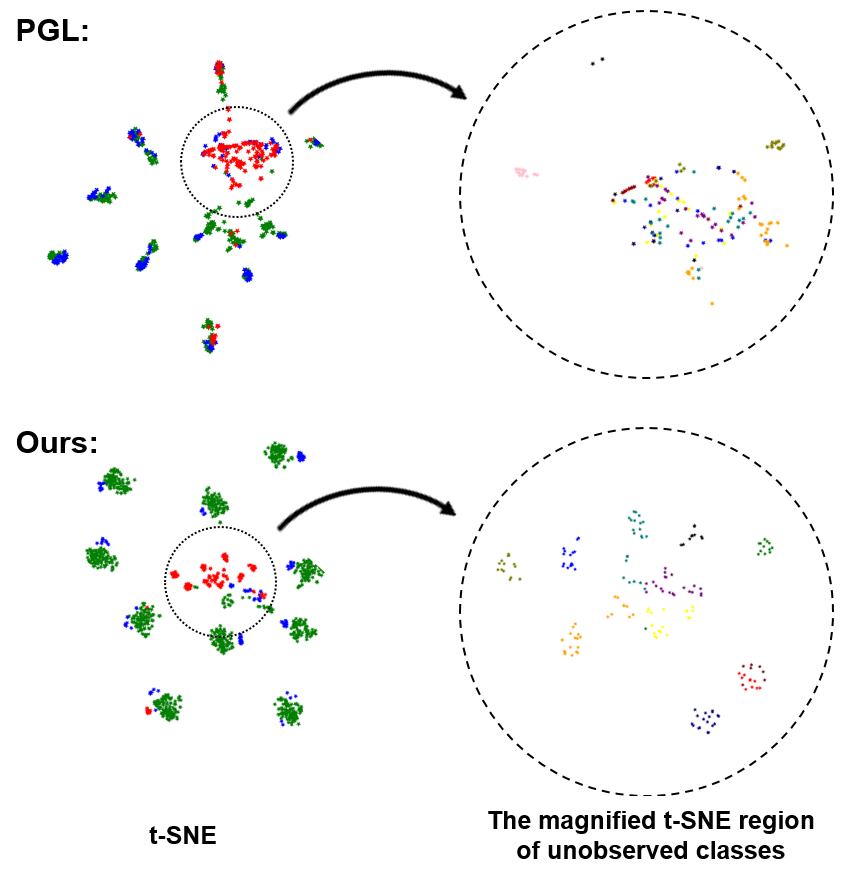}
	\caption{(\textbf{Left}).T-SNE visualizations of the features learned by PGL and SCDA on task A$\rightarrow$D in OSDA setup. Green points indicate source features, blue points indicate target known features, and red points indicate target unknown features. (\textbf{Right}).We magnify the region that contains the features of unobserved classes and use different colors separate their categories.
	}
	\label{pic_vis}       % Give a unique label
\end{figure}

\subsection{Analysis}\label{analysis}			
\textbf{The ratio of unobserved classes.} The traditional OSDA benchmarks usually fix the ratio of unobserved classes into 0.5. But in real-world tasks, we are not aware of the ratio which can vary drastically. To validate the robustness to the changing ratio, we conduct the experiments on \textbf{Office-31} with variant ratios. As illustrated in Fig \ref{pic_ratio1}, no matter how the ratio changes, SCDA performs stably and precisely estimates the unobserved class number. Furthermore, we compare SCDA with STA and OSBP in task A$\rightarrow$W on \textbf{Office-31} by varying the ratio. As shown in Fig \ref{pic_ratio2}, all baselines preserve consistent OS* regardless of the changing ratio. But the accuracies of OSBP and STA descend sharply as the unobserved class number increases. In a comparison, SCDA achieves the more stable performance in OS* along with the persistent accuracy of unknown samples. It reveals that SCDA is insensitive to the change of unobserved class number.

\textbf{Visualization.} For the task A$\rightarrow $D in \textbf{Office-31}, we visualize the last-layer features from PGL, and SCDA by t-SNE. In Figures \ref{pic_vis}, PGL tends to push unknown target features together. However, the groups of unknown features are mixed with the known-classes samples. It suggests that unifying all unobserved classes into ``\emph{unknown}'' causes confusing decision boundaries that decrease the OSDA performance. While the features generated from SCDA present significant classification margins across domain-shared and unobserved classes. Further, we magnify the region of the unknown features and use different colors to indicate different unknown categories. As shown in the dotted circles, SCDA features present more significant margins between unobservable categories whereas STA simply mixes them together. The illustration verifies the hypothesis in our introduction and demonstrates SCDA's capability to discover unseen classes in target samples.


\end{document}